\documentclass[10pt,twocolumn,letterpaper]{article}

\usepackage{iccv}
\usepackage{times}
\usepackage{epsfig}
\usepackage{graphicx}
\usepackage{amsmath}
\usepackage{amssymb}

\usepackage{algorithm}
\usepackage{algorithmic}

\newcommand{\norm}[1]{\left\lVert#1\right\rVert}
\usepackage{multirow}
\usepackage{verbatim}

\usepackage[breaklinks=true,bookmarks=false]{hyperref}

\iccvfinalcopy 

\def\iccvPaperID{****} 

\begin{document}

\title{STD: Sparse-to-Dense 3D Object Detector for Point Cloud}

\author{Zetong Yang$^{\dag}$\\
\and
Yanan Sun$^{\dag}$
\and
Shu Liu$^{\dag}$
\and 
Xiaoyong Shen$^{\dag}$
\and 
Jiaya Jia$^{\dag, \ddagger}$
\\
\and
$^{\dag}$Youtu Lab, Tencent~~~~~~$^{\ddagger}$The Chinese University of Hong Kong\\
\vspace{-2mm}
{\tt\small \{tomztyang, now.syn, liushuhust, Goodshenxy\}@gmail.com~~leojia@cse.cuhk.edu.hk} 
}
\maketitle

\begin{abstract}
   We present a new two-stage 3D object detection framework, named sparse-to-dense 3D Object Detector (STD). The first stage is a bottom-up proposal generation network that uses raw point cloud as input to generate accurate proposals by seeding each point with a new spherical anchor. It achieves a high recall with less computation compared with prior works. Then, PointsPool is applied for generating proposal features by transforming their interior point features from sparse expression to compact representation, which saves even more computation time. In box prediction, which is the second stage,  we implement a parallel intersection-over-union (IoU) branch to increase awareness of localization accuracy, resulting in further improved performance. We conduct experiments on KITTI dataset, and evaluate our method in terms of 3D object and Bird's Eye View (BEV) detection. Our method outperforms other state-of-the-arts by a large margin, especially on the hard set, with inference speed more than 10 FPS. 
\end{abstract}

\section{Introduction}
3D scene understanding with point cloud is a very important topic in computer vision, since it benefits many applications, such as autonomous driving \cite{KITTIDATASET2} and augmented reality \cite{Multiple3Dtracking}. In this work, we focus on one essential 3D scene recognition task, object detection based on point cloud, which predicts the 3D bounding box and class label for each object in the scene.

Compared to RGB images, LiDAR point cloud has its own unique properties. On the one hand, they provide structural and spatial information of relative location and precise depth. On the other hand, they are unordered, sparse and locality sensitive, which brings difficulties in parsing raw LiDAR point cloud.

Most existing work transforms sparse point clouds to compact representations by projecting it to images \cite{MV3D,AVOD,MultiViewRandomForest,PedestrianDetectionCombine,Vote3Deep} or subdividing it into equally distributed voxels \cite{VoxNet,VotetoVote,VOXELNET,YangLU18}. CNNs can be applied for parsing the point cloud. Nevertheless, these hand-crafted representations may not be optimal. 
Instead of converting irregular point cloud to voxels, Qi \etal proposes PointNet \cite{POINTNET,POINTNET2} to directly operate on raw LiDAR point clouds for classification and semantic segmentation. Two streams of methods tackling 3D object detection follow. One is based on voxels, \eg, VoxelNet \cite{VOXELNET} and SECOND \cite{yan2018second}, where voxelization is conducted on the entire point cloud, PointNet is applied to each voxel for feature extraction and CNNs are used for final bounding-box prediction. Although efficient, information loss is inevitable, degrading localization quality. The other stream is point-based, like F-PointNet \cite{FPOINTNET} and PointRCNN \cite{shi2018pointrcnn}. They take raw point cloud data as input, and generate final predictions by PointNet++ \cite{POINTNET2}. These methods achieve better performance with uncontrollable receptive fields and large computation cost.

\begin{figure*}[t]
	\centering
	\includegraphics[width=1.0\linewidth]{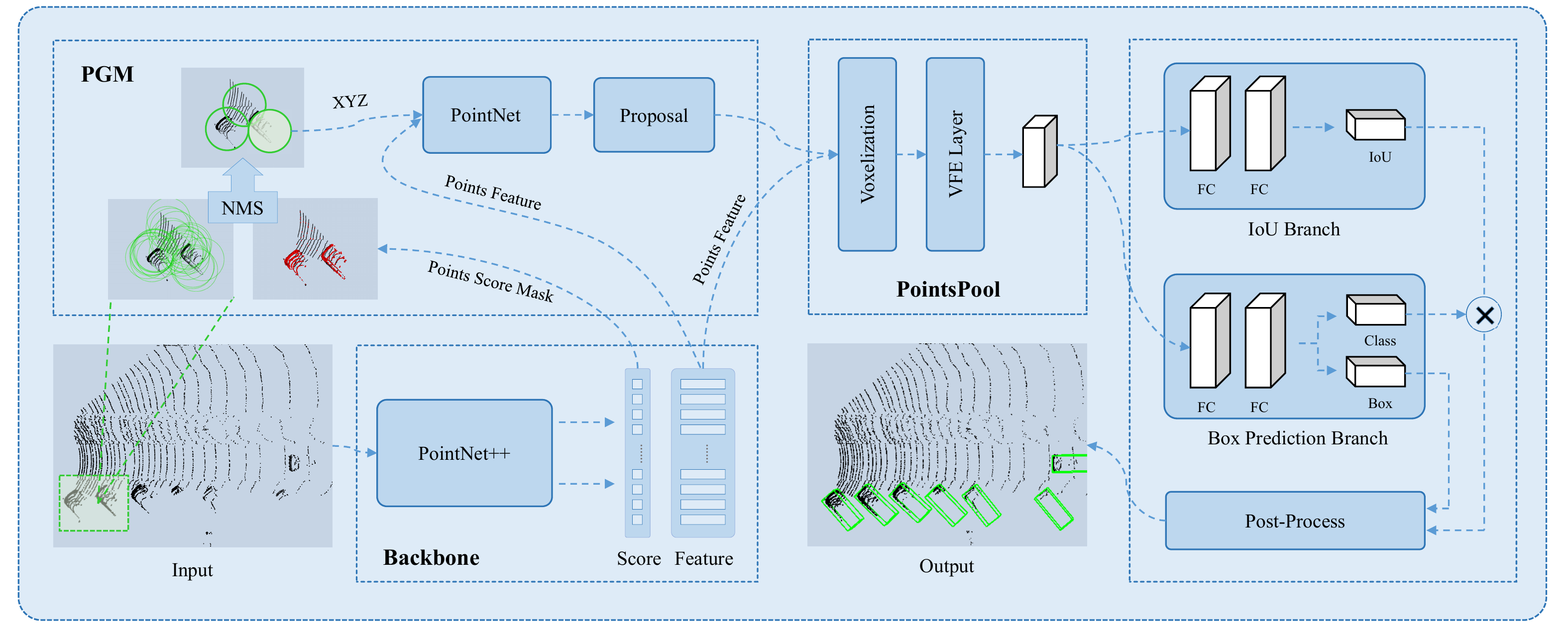}\\
	\caption{Illustration of our framework consisting of three different parts. The first is a proposal generation module (PGM) to generate  accurate proposals from man-made point-based spherical anchors. The second part is a PointsPool layer to convert proposal features from sparse expression to compact representation. The final one is a box prediction network. It classifies and regresses proposals, and picks high-quality predictions.}
	\label{fig:framework}\vspace{-0.1in}
\end{figure*}

\vspace{-0.1in}
\paragraph{Our Contributions}
Different from all previous methods, we propose a two-stage 3D object detection framework. In the first stage, we take each point in the point cloud as an element, and seed them with appropriate spherical anchors, aiming to preserve accurate location information. Then a PointNet++ backbone is applied for extracting semantic context feature for each point as well as generating objectness score to filter anchors. 

To generate feature for each proposal, we propose the PointsPool layer by gathering canonical coordinates and semantic features of their interior points, retaining accurate localization and context information. This layer transforms sparse and unordered point-wise expression to more compact features, enabling utilization of efficient CNNs and end-to-end training. Final prediction is achieved in the second stage. Instead of predicting the box location and class label with a simple head, we propose to augment a novel 3D IoU branch for predicting 3D IoU between predictions and ground-truth bounding boxes to alleviate inappropriate removal during post-processing.

We evaluate our model on KITTI dataset \cite{KITTI3DBENCHMARK}. Experiments show that our model outperforms other state-of-the-arts in terms of both BEV and 3D object detection tasks, especially for difficult examples. Our primary contribution is manifold.

\begin{itemize}\vspace{-0.05in}
\item We propose a point-based proposal generation paradigm for object detection on point cloud with spherical anchors. It is generic to achieve high recall. \vspace{-0.05in}

\item The proposed PointsPool layer integrates advantages of both point- and voxel-based methods, enabling efficient and effective prediction.
\vspace{-0.05in}

\item Our new 3D IoU prediction branch helps alignment between classification score and localization, leading to notable improvement. Experimental results on KITTI dataset show that our framework handles many challenging cases with high occlusion and crowdedness, and achieves new state-of-the-art performance. Moreover, with our design, the good performance is achieved at a 10 FPS speed.
\end{itemize}

\section{Related Work}

\paragraph{3D Semantic Segmentation} There are several approaches to tackle semantic segmentation on point cloud. In \cite{SQUEEZESEG}, a projection function converts LiDAR points to a UV map, which is then classified by 2D semantic segmentation \cite{SQUEEZESEG,PSPNET,DEEPLAB} in pixel level. In \cite{3DMV,SCANNET}, a multi-view-based function produces the segmentation mask. This method fuses information from different views. Other solutions, such as \cite{POINTNET2,POINTNET,POINTCNN,POINTSIFT,SONET}, segment point cloud from raw LiDAR data. They directly generate features on each point while keeping original structural information. A max-pooling method gathers the global feature. It is then concatenated with local feature for processing.

\vspace{-0.05in}
\paragraph{3D Object Detection} There are three different lines for 3D object detection. They are multi-view, voxel, and point-based methods.

For multi-view methods, MV3D \cite{MV3D} projects LiDAR point cloud to BEV and trains a Region Proposal Network (RPN) to generate positive proposals. It merges features from BEV, image view and front view in order to generate refined 3D bounding boxes. AVOD \cite{AVOD} improves MV3D by fusing image and BEV features like \cite{FPN}. Unlike MV3D, which only merges features in the refinement phase, it also merges features from multiple views in the RPN phase to generate positive proposals. These methods still have the limitation when detecting small objects such as pedestrians and cyclists. They do not deal with cases with multiple objects in depth direction.

There are several LiDAR-data based 3D object detection frameworks using voxel-grid representation. In \cite{VotetoVote}, each non-empty voxel is encoded with 6 statistical quantities by the points within this voxel. Binary encoding is used in \cite{FullyConvolutionNetworkForVehicle} for each voxel grid. In PIXOR \cite{YangLU18}, each voxel grid is encoded as occupancy.  All of these methods use hand-crafted representation. VoxelNet \cite{VOXELNET} instead stacks many VFE layers to generate machine-learned representation for each voxel. Compared to \cite{VOXELNET}, SECOND \cite{yan2018second} uses sparse convolution layers \cite{sparseconv} for parsing the compact representation. PointPillars \cite{lang2018pointpillars} uses pseudo-images as the representation after voxelization.

F-PointNet \cite{FPOINTNET} is the first method of utilizing raw point cloud to predict 3D objects. It uses frustum proposals from 2D object detection as candidate boxes and regresses predictions based on interior points. Therefore, performance heavily relies on the 2D object detector. Differently, PointRCNN \cite{shi2018pointrcnn} uses the whole point cloud for proposal generation rather than 2D images. It directly uses the segmentation score of proposal's centric point for classification considering proposal location information. Other features like size and orientation are neglected. In contrast, our design is general to utilize the strong representation power of point cloud. 

\begin{figure*}[bpt]
	\centering
	\includegraphics[width=1.0\linewidth]{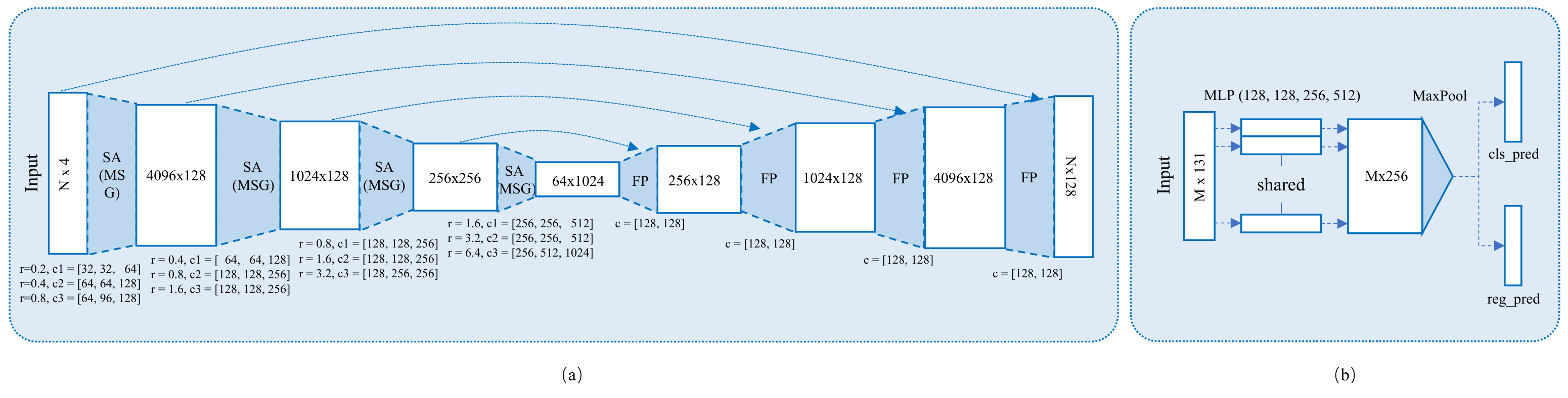}\\\vspace{-0.1in}
	\caption{Illustration of networks in the proposal generation module. (a) 3D segmentation network (PointNet++). It takes a raw point cloud $(x, y, z, r)$ as input, and generates semantic segmentation scores as well as global context features for each point by stacking SA layers and FP modules. (b) Proposal generation Network (PointNet). It treats normalized coordinates and semantic features of points within anchors as input, and produces classification and regression predictions.}
	\label{fig:networks}
\end{figure*}

\section{Our Framework}

Our method is a two-stage 3D object detection framework exploiting advantages of voxel- and point-based methods. To generate accurate point-based proposals, we design spherical anchors and a new strategy in assigning labels to anchors. For each generated proposal, we deploy a new PointsPool layer to convert point-based features from sparse expression to dense representation. A box prediction network is applied for final prediction. The framework is illustrated in Figure \ref{fig:framework}.  

\subsection{Proposal Generation Module}

Existing methods of 3D object detection mainly project point cloud to different views or divide them into voxels for utilizing CNNs. We instead design a generic strategy to seed anchors based on each point independently, which is the elementary component in the point cloud. Then features of interior points of each anchor are utilized to generate proposals. With this structure, we keep sufficient context information, achieving decent recall even with a small number of proposals.

\vspace{-0.1in}
\paragraph{Challenge} Albeit elegant, point-based frameworks inevitably face many challenges. For example, the amount of points is prohibitively huge where high redundancy exists in anchors. They cost much computation during training and inference. Also, the way to assign ground-truth labels for anchors needs to be specially designed.

\vspace{-0.1in}
\paragraph{Spherical Anchor} The first step of proposal generation module is to reasonably seed anchors for each point. Considering that a 3D object could be with any orientations, we design spherical anchors rather than traditional cuboid anchors. For each spherical anchor, it is with a spherical receptive field parametrized by the class-specific radius (\ie, 2-meter radius for car, and 1-meter radius for pedestrian and cyclist). Now the proposal predicted by each anchor is based on the points in the spherical receptive field. Each anchor is associated with a reference box for proposal generation, with pre-defined size. These anchors are located at the center of each point. Different from traditional anchor schemes, we do not pre-define the orientation of the reference box. It is instead directly predicted. As a result, the number of spherical anchors is not proportional to the number of pre-defined reference box orientation, leading to about 50\% less anchors. With computation much reduced, we surprisingly achieve a much higher recall with spherical anchors than with traditional ones.

This step reduces the amount of anchors to about 16K. To further compress them, we use a 3D semantic segmentation network to predict the class of each point and produce semantic feature for each point. It is followed by non-maximal suppression (NMS) to remove redundant anchors.
The final score of each anchor is the segmentation score on the center point. The IoU value is calculated based on the projection of each anchor to the BEV. With these operations, we reduce the number of anchors to around only 500.

\vspace{-0.1in}
\paragraph{Proposal Generation Network}
\label{paragraph_proposla_generation_network}
These computed useful anchors lead to accurate proposals. Inspired by PointNet \cite{POINTNET} in 3D classification, we gather 3D points within anchors for regression and classification. For points in an anchor, we pass their $(X,Y,Z)$ locations, which are normalized by the anchor center coordinates, and semantic features from the segmentation network to a PointNet with several convolutional layers to predict classification scores, regression offsets and orientations. Details of the 3D segmentation networks and PointNet are illustrated in Figure \ref{fig:networks}. 

Then we compute offsets regarding anchor center coordinates $(A_x, A_y, A_z)$ and their pre-defined sizes $(A_l, A_w, A_h)$ so as to obtain precise proposals. 
The pre-defined size for ``car", ``cyclist" and ``pedestrian" are $(A_l=3.9, A_w=1.6, A_h=1.6)$, $(A_l=1.6, A_w=0.8, A_h=1.6)$ and $(A_l=0.8, A_w=0.8, A_h=1.6)$, respectively. 
For angle prediction, we use a hybrid of classification and regression formulation following \cite{FPOINTNET}. That is, we pre-define $N_a$ as equally split angle bins and classify the proposal angle into different bins. Residual is regressed with respect to the bin value. $N_a$ is set to 12 in our experiments. Finally, we apply NMS based on classification score and oriented BEV IoU to eliminate redundant proposals. Specifically, we keep up to 300 proposals during training and 100 for testing.

\vspace{-0.1in}
\paragraph{Assignment Strategy}
Given that our anchors are with spherical receptive fields rather than cubes or cuboids, it is not appropriate to assign positive or negative labels according to traditional IoU calculation \cite{VOXELNET} between the spherical receptive field and ground-truth boxes. We design a new criterion named PointsIoU to assign target labels. PointsIoU is defined as the quotient between the number of points in the intersection area of both regions and the number of points in the union area of both regions.
An anchor is considered positive if its PointsIoU with a certain ground-truth box is higher than 0.55 and negative otherwise.

\subsection{Proposal Feature Generation}
With semantic features from the segmentation network for each point and refined proposals, we constitute compact features for each proposal. 

\vspace{-0.1in}
\paragraph{Motivation}
For each proposal, the most straight-forward way to make final prediction is to perform PointNet++ based on interior points \cite{shi2018pointrcnn,FPOINTNET}. Albeit simple, several operations such as {\it set\ abstraction} (SA) are computational expensive compared to traditional convolution or fully connected (FC) layers. 
As illustrated in Table \ref{tab:whether_pointspool}, with 100 proposals, PointNet++ baseline takes 41ms during inference, compared with 16ms with pure FC layers. It is almost $2.5 \times$ faster than the baseline, with only $0.4\%$ performance drop. Moreover, compared to PointNet baseline, the model with FC layers yields $1.6 \%$ performance increase with only 6 extra milliseconds. It is because PointNet regression head uses less local information. 

We apply a voxelization layer at this stage, named PointsPool, to compute compact proposal features that can be used in efficient FC layers for final predictions. Compared to voxelization in \cite{VOXELNET}, this new layer is a gradient-conductive voxelization layer, enabling end-to-end training.


\vspace{-0.1in}
\paragraph{PointsPool Layer}
PointsPool layer is composed of three steps. In the first step, we randomly choose $N$ interior points for each proposal with their canonical coordinates and semantic features as initial feature. For each proposal, we obtain point canonical locations by subtracting the proposal center $(X,Y,Z)$ values and rotating them to the proposal predicted orientation. These canonized coordinates enable the model to be robust under geometrical transformation, and be aware of inner points' relative locations for better performance than only using semantic features.

The second step is using the voxelization layer to subdivide each proposal into equally spaced voxels as \cite{VOXELNET}. Specifically, we partition each proposal to $(d_l=6, d_w=6, d_h=6)$ voxels. $N_r=35$ points are randomly sampled for each voxel. Concatenated features of canonical coordinates and semantic features of these points are used for each voxel. Compared to voxelization in \cite{VOXELNET}, this layer has gradient representation, making it possible for end-to-end training. While passing gradients, we only need to pass gradients of these randomly selected points. Finally, we apply a Voxel Feature Encoding (VFE) layer with channels $(128, 128, 256)$ \cite{VOXELNET} for extracting features of each voxel, so as to generate features of proposals with shapes $(d_l \times d_w \times d_h \times 256)$. 
After getting features of each proposal, we flatten them for following FC layers in the box prediction head. 

\begin{table}[t]
   \centering \addtolength{\tabcolsep}{-1pt}
   \footnotesize
   \begin{tabular}{|c|c|c|c|}
       \hline
       Methods & Proposal No. & Inference Time & Moderate \\
       \hline
       PointNet (4 conv layers) & 100 & 10 ms& 77.1  \\
       PointNet++ (3 SA) & 100 & 41 ms & 79.1 \\
       PointsPool + 2FC & 100 & 16 ms & 78.7 \\
      \hline
   \end{tabular}\vspace{0.3cm}
   \caption{3D object detection AP on KITTI val moderate set. We compare inference time and AP among different box regression network architectures.}
   \label{tab:whether_pointspool}
\end{table} 

\subsection{Box Prediction Network}
Our box prediction network has two branches for box estimation and IoU estimation respectively.

\vspace{-0.1in}
\paragraph{Box Estimation Branch}
In this branch, we use 2 FC layers with channels $(512, 512)$ to extract features of each proposal. Then another 2 FC layers are applied for classification and regression respectively. We directly regress offsets between the ground-truth box and proposals, parametrized by $(t_l, t_w, t_h)$. We further predict the shift $(t_x, t_y, t_z)$ from proposal center to the ground-truth. As for angle prediction, we still use a hybrid of classification and regression formulation, same as the one described in Section \ref{paragraph_proposla_generation_network}.

\begin{table}[t]
   \centering \addtolength{\tabcolsep}{-1pt}
   \footnotesize
   \begin{tabular}{|c|c|c|c|c|c|}
       \hline
       Score-NMS & IoU-NMS & Easy & Moderate & Hard \\
       \hline
       $\surd$ & - & 88.8 & 78.7 & 78.2 \\
       - & $\surd$ & 90.9 & 90.9 & 90.6 \\
      \hline
   \end{tabular}\vspace{0.3cm}
   \caption{3D object detection AP on KITTI val set. We conduct experiments to show importance of the post process. ``Score-NMS'' means using classification scores as NMS sorting scores. ``IoU-NMS'' means for each prediction, we use its largest IoU among all ground-truth boxes as the sorting score. }
   \label{tab:whether_nms_iou_nearest}
\end{table}

\vspace{-0.1in}
\paragraph{IoU Estimation Branch}
\label{IoU_estimation_framework}
In previous work \cite{lang2018pointpillars,yan2018second,VOXELNET,AVOD,shi2018pointrcnn}, NMS is applied to results of box estimation to remove duplicate predictions. The classification score is used for ranking during NMS. Noted in \cite{jiang2018acquisition,liu2015box,qi2018sequential}, the classification scores of boxes are not highly correlated with the localization quality. Similarly, weak correlation between classification score and box quality affects point-based object detection tasks. Given that LiDAR for autonomous driving is usually gathered at a fixed angle, and objects are partially covered, localization accuracy is extremely sensitive to relative position between visible part and its full view while the classification branch cannot provide enough information. 
As shown in Table \ref{tab:whether_nms_iou_nearest}, if we feed the oracle IoU value of each predicted box instead of the classification score to NMS for duplicate removal, the performance increases by around $12.6 \%$.

Based on this fact, we develop an IoU estimation branch for predicting 3D IoU between boxes and corresponding ground-truth. Then, we multiply each box's classification score with its 3D IoU as a new sorting criterion. This design relieves the discrepancy between localization accuracy and classification score, effectively improving the final performance. Moreover, this IoU estimation branch is general and can be applied to other 3D object detectors. We expect similar performance improvement on other frameworks.

\subsection{Loss Function}
We use a multi-task loss to train our network. Our total loss is composed of proposal generation loss $L_{prop}$ and box prediction loss $L_{box}$ as 
\begin{equation} \label{eq:total_loss}
\begin{aligned}
L_{total} &= L_{prop} + L_{box}.
\end{aligned}
\end{equation}
The proposal generation loss is the summation of 3D semantic segmentation loss and proposal prediction loss. We use focal loss \cite{FocalLoss} as segmentation loss $L_{seg}$, keeping the original parameters $\alpha_t = 0.25$ and $\gamma = 2$. The prediction loss consists of proposal classification loss and regression loss. The overall proposal generation loss is defined as Eq. \eqref{eq:proposal_loss}. $s_i$ and $u_i$ are the predicted classification score and ground-truth label for anchor $i$, respectively. $N_{cls}$ and $N_{pos}$ are the number of anchors and positive samples.
\begin{equation} \label{eq:proposal_loss}
\begin{aligned}
L_{prop} &= L_{seg} + \frac{1}{N_{cls}} \sum_i L_{cls} (s_i, u_i) \\
&+ \lambda \frac{1}{N_{pos}} \sum_i  [u_i \geq 1] (L_{loc} + L_{ang}),
\end{aligned}
\end{equation}
where the Iverson bracket indicator function $[u_i \geq 1]$ reaches 1 when $u_i \geq 1$ and 0 otherwise. 
$L_{cls}$ is simply the softmax cross-entropy loss.
We parameterize an anchor $A$ by its center $(A_x, A_y, A_z)$ and size $(A_l, A_w, A_h)$. 
Its ground truth box $G$ is with $(G_x, G_y, G_z)$ and $(G_l, G_w, G_h)$.
Location regression loss is composed of center residual prediction loss and size residual prediction loss, formulated as
\begin{equation} \label{eq:regnoangle}
\begin{aligned}
L_{loc} =\ L_{dis}(A_{ctr}, G_{ctr}) + L_{dis}(A_{size}, G_{size}),
\end{aligned}
\end{equation}
where $L_{dis}$ is the smooth-$l_1$ loss. $A_{ctr}$ and $A_{size}$ are predicted center residual and size residual by the proposal generation network, while $G_{ctr}$ and $G_{size}$ are targets for them. The target of our network is defined as
\begin{eqnarray} 
\left \{ \begin{array}{llll}
        G_{ctr}&=& G_j - A_j \textrm{,} & j \in (x,\ y,\ z)\\
        G_{size} &=& (G_j - A_j)/A_j \textrm{,} & j \in (l,\ w,\ h).\\
        \end{array} \right.
\end{eqnarray}
Angle loss includes orientation classification loss and residual prediction loss as
\begin{equation} \label{eq:anglereg}
L_{angle} = L_{cls}(t_{a-cls}, v_{a-cls}) + L_{dis}(t_{a-res}, v_{a-res}),
\end{equation}
where $t_{a-cls}$ and $t_{a-res}$ are predicted angle class and residual while $v_{a-cls}$ and $v_{a-res}$ are their targets. 

The box prediction loss is almost the same as the proposal prediction loss, mentioned above, with two extra losses, which are 3D IoU loss and corner loss. When training IoU branch, we use 3D IoU between proposals and corresponding ground-truth boxes as ground-truth, and smooth-$l_1$ loss as the loss function. Corner loss is the distance between the predicted 8 corners and assigned ground-truth, expressed as
\begin{equation} \label{eq:cornerloss}
L_{corner} = \sum_{k=1}^8 \norm{P_{k} - G_{k}},
\end{equation}
where $P_{k}$ and $G_{k}$ are the location of ground-truth and prediction for point $k$.  

\begin{table*}[t]
   \centering 
   \footnotesize
   \begin{tabular}{|c|c|c||c|c|c||c|c|c|}
       \hline
       \multicolumn{1}{|c|}{ \multirow{2}{*}{Class}} & \multicolumn{1}{c|}{ \multirow{2}{*}{Method}} & \multicolumn{1}{c||}{ \multirow{2}{*}{Modality}} & \multicolumn{3}{|c||}{$AP_{BEV} (\%)$}& \multicolumn{3}{|c|}{$AP_{3D} (\%)$} \\ \cline{4-9}
       \multicolumn{1}{|c|}{} & \multicolumn{1}{c|}{} & \multicolumn{1}{c||}{} & \multicolumn{1}{|c|}{Easy} & \multicolumn{1}{|c|}{Moderate} & \multicolumn{1}{|c||}{Hard} & \multicolumn{1}{|c|}{Easy} & \multicolumn{1}{|c|}{Moderate} & \multicolumn{1}{|c|}{Hard} \\
       \hline
       \hline
      \multirow {11}{*}{Car} & MV3D \cite{MV3D} & \multirow{6}{*}{RGB + LiDAR}  & 86.02 & 76.90 & 68.49 & 71.09 & 62.35 & 55.12\\
      {} & AVOD \cite{AVOD} & {} & 86.80 &  85.44 & 77.73 & 73.59 & 65.78 & 58.38 \\
      {} & F-PointNet \cite{FPOINTNET} & {} & 88.70 & 84.00 & 75.33 & 81.20 & 70.39 & 62.19 \\ 
      {} & AVOD-FPN \cite{AVOD} & {} & 88.53 & 83.79 &  77.90 &  81.94 & 71.88 &  66.38 \\
      {} & RoarNet \cite{kiwoo2018roar} & {} & 88.75 & 86.08 & 78.80 & 83.95 & 75.79 & 67.88 \\
      {} & UberATG-MMF \cite{Liang2019CVPR} & {} & 89.49 & 87.47 & 79.10 & \bf 86.81 & 76.75 & 68.41 \\ \cline{2-9}
      {} & VoxelNet \cite{VOXELNET} & \multirow{5}{*}{LiDAR} &  89.35 & 79.26 & 77.39 & 77.47 & 65.11 & 57.73 \\
      {} & SECOND \cite{yan2018second} & {} & 88.07  & 79.37 & 77.95 & 83.13 & 73.66 & 66.20 \\
      {} & PointPillars \cite{lang2018pointpillars} & {} & 88.35 & 86.10 & 79.83 & 79.05 & 74.99 & 68.30 \\
      {} & PointRCNN \cite{shi2018pointrcnn} & {} & 89.47 & 85.68 & 79.10 & 85.94 & 75.76 & 68.32 \\
      {} & Ours & {} & \bf 89.66 & \bf 87.76 & \bf 86.89 & 86.61 & \bf 77.63 & \bf 76.06 \\
      \hline
      \hline
      \multirow {8}{*}{Pedestrian} & AVOD \cite{AVOD} & \multirow{3}{*}{RGB + LiDAR} & 42.51 & 35.24 & 33.97 & 38.28 & 31.51 & 26.98 \\
      {} & F-PointNet \cite{FPOINTNET} & {} & 58.09 & 50.22 & 47.20 & 51.21 & \bf 44.89 & 40.23 \\ 
      {} & AVOD-FPN \cite{AVOD} & {} & 58.75 & 51.05 & \bf 47.54 & 50.80 & 42.81 & 40.88 \\ \cline{2-9}
      {} & VoxelNet \cite{VOXELNET} & \multirow{5}{*}{LiDAR} & 46.13 & 40.74 & 38.11 & 39.48 & 33.69 & 31.51 \\
      {} & SECOND \cite{yan2018second} & {} & 55.10 & 46.27 & 44.76 & 51.07 & 42.56 & 37.29 \\
      {} & PointPillars \cite{lang2018pointpillars} & {} & 58.66 & 50.23 & 47.19 & 52.08 & 43.53 & 41.49 \\
      {} & Ours & {} & \bf 60.99 & \bf 51.39 & 45.89 & \bf 53.08 & 44.24 & \bf 41.97 \\
      \hline
      \hline
      \multirow {8}{*}{Cyclist} & AVOD \cite{AVOD}  & \multirow{3}{*}{RGB + LiDAR} & 63.66 & 47.74 & 46.55 & 60.11 & 44.90 & 38.80 \\
      {} & F-PointNet \cite{FPOINTNET} & {} & 75.38 & 61.96 & 54.68 & 71.96 & 56.77 & 50.39 \\ 
      {} & AVOD-FPN \cite{AVOD} & {} & 68.09 & 57.48 & 50.77 & 64.00 & 52.18 & 46.61 \\ \cline{2-9}
      {} & VoxelNet \cite{VOXELNET} & \multirow{5}{*}{LiDAR} & 66.70 & 54.76 & 50.55 & 61.22 & 48.36 & 44.37 \\
      {} & SECOND \cite{yan2018second} & {} & 73.67 & 56.04 & 48.78 & 70.51 & 53.85 & 46.90 \\
      {} & PointPillars \cite{lang2018pointpillars} & {} & 79.14 & 62.25 & 56.00 & 75.78 & 59.07 & 52.92 \\
      {} & Ours & {} & \bf 81.04 & \bf 65.32 & \bf 57.85 & \bf 78.89 & \bf 62.53 & \bf 55.77 \\
      \hline
   \end{tabular}\vspace{0.1cm}
   \caption{Performance on KITTI test set for both Car, Pedestrian and Cyclists.\vspace{-0.1in}}\label{tab:mainkitti}
\end{table*}

\section{Experiments}
We evaluate our method on the widely used KITTI Object Detection Benchmark \cite{KITTI3DBENCHMARK}. There are 7,481 training images / point clouds and 7,518 test images / point clouds with three categories of Car, Pedestrian and Cyclist. We use average precision (AP) metric to compare with different methods. During evaluation, we follow the official KITTI evaluation protocol -- that is, the IoU threshold is 0.7 for class Car and 0.5 for Pedestrian and Cyclist. 

\subsection{Implementation Details}
Following former works \cite{VOXELNET,lang2018pointpillars,AVOD,yan2018second}, in order to avoid IoU misalignment in KITTI evaluation protocol on Car, Pedestrian and Cyclist, we train two networks, one for car and the other for both pedestrian and cyclist.

\vspace{-0.1in}
\paragraph{Network Architecture}
To align network input, we randomly choose 16K points from the entire point cloud for each scene.
Our 3D semantic segmentation network is based on PointNet++ with four SA levels and four {\it feature propagation} (FP) layers. 
The proposal generation sub-network is a multi-layer perception consisting of four hidden layers with channels $(128, 128, 256, 512)$, followed by a PointsPool layer where we randomly sample $N=512$ interior points per proposal as its initial input. These representations are then passed to the box regression network. Both the box estimation and IoU estimation branches consist of 2 fully connected layers with 512 channels. 

\vspace{-0.1in}
\paragraph{Training Parameters} 
Our model is trained stage-by-stage to save GPU memory. The first stage consists of 3D semantic segmentation and proposal generation, while the second is for box prediction. For the first stage, we use ADAM \cite{AdamOptimizer} optimizer with an initial learning rate 0.001 for the first 80 epochs and then decay it to 0.0001 for the last 20 epochs. Each batch consists of 16 point clouds evenly distributed on 4 GPU cards. 
For the second stage, we train 50 epochs with batch size 1. The learning rate is initialized as 0.001 for first 40 epochs and is then decayed by 0.1 in every 5 epochs.
For each input point cloud, we sample 256 proposals, with ratio 1:1 for positives and negatives.
Our implementation is based on Tensorflow \cite{Tensorflow}. For the box prediction network, a proposal is considered positive if its maximum 3D IoU with all ground-truth boxes is higher than 0.55 and negative if its maximum 3D IoU is below 0.45 during training the car model. The positive and negative 3D IoU thresholds are 0.5 and 0.4 for the pedestrian and cyclist models. 
Besides, for the IoU branch, we only train on positive proposals.

\vspace{-0.1in}
\paragraph{Data Augmentation}
Data augmentation is important to prevent overfitting. First, similar to that of \cite{yan2018second}, we randomly add several ground-truth boxes with their interior points from other scenes to current point cloud in order to simulate objects with various environments. Then, for each bounding box, we randomly rotate it following a uniform distribution $\Delta \theta_1 \in [- \pi / 4, + \pi / 4]$ and randomly add a translation ($\Delta x, \Delta y, \Delta z$). Third, each point cloud is flipped along the $x$-axis in camera coordinate with probability 0.5. We also randomly rotate each point cloud around $z$-axis (up axis) 
by a uniformly distributed random variable $\Delta \theta_2 \in [- \pi / 4, + \pi / 4]$. Finally, we apply a global scaling to point cloud with a random variable drawn from the uniform distribution $[0.9, 1.1]$.

\subsection{Main Results}

For evaluation on the test set, we train the model on split train/val sets at ratio 4:1.
The performance of our method and comparison with previous methods are listed in Table \ref{tab:mainkitti}. Our model outperforms other methods by a large margin on Car and Cyclist classes, especially on the hard set. Compared to multi-view methods that use other sensors as extra information, our method still achieves higher AP with input of only raw point cloud. Compared to UberATG-MMF \cite{Liang2019CVPR}, which is the best multi-sensor detector, STD outperforms it by $0.88 \%$ on the moderate level on 3D detection of Cars. Also large increase $7.65 \%$ on the hard set is obtained, manifesting the effectiveness of our proposal-generation module and IoU branch.

\begin{figure}[bpt]
	\centering
	\includegraphics[width=0.95\linewidth]{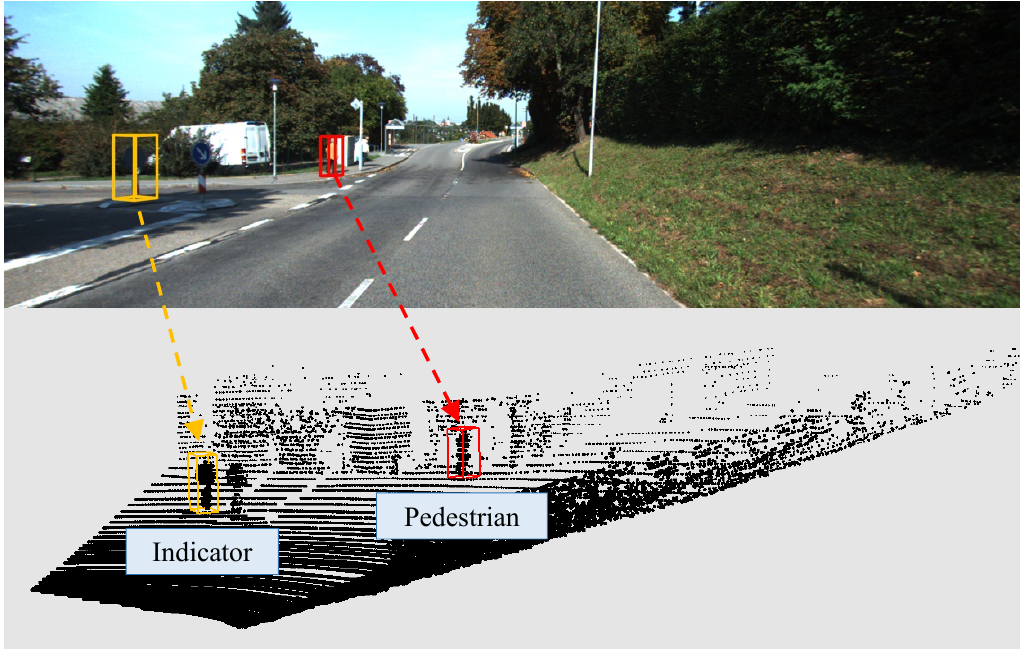}
	\caption{Small objects such as indicators are easy to detect on RGB images, but not on LiDAR data.}
	\label{fig:inferior_pedestrian}
\end{figure}

Note that on Pedestrian class, STD is still the best among LiDAR-only detectors. Multi-sensor detectors work better because there are very few 3D points on pedestrians, making it difficult to distinguish them from other small objects like indicator or telegraph pole, as shown in Figure \ref{fig:inferior_pedestrian}. Extra information of RGB would help in these cases.

\begin{figure*}[t]
	\centering
	\begin{tabular}{@{\hspace{0mm}}c}
		\includegraphics[width=0.95\linewidth]{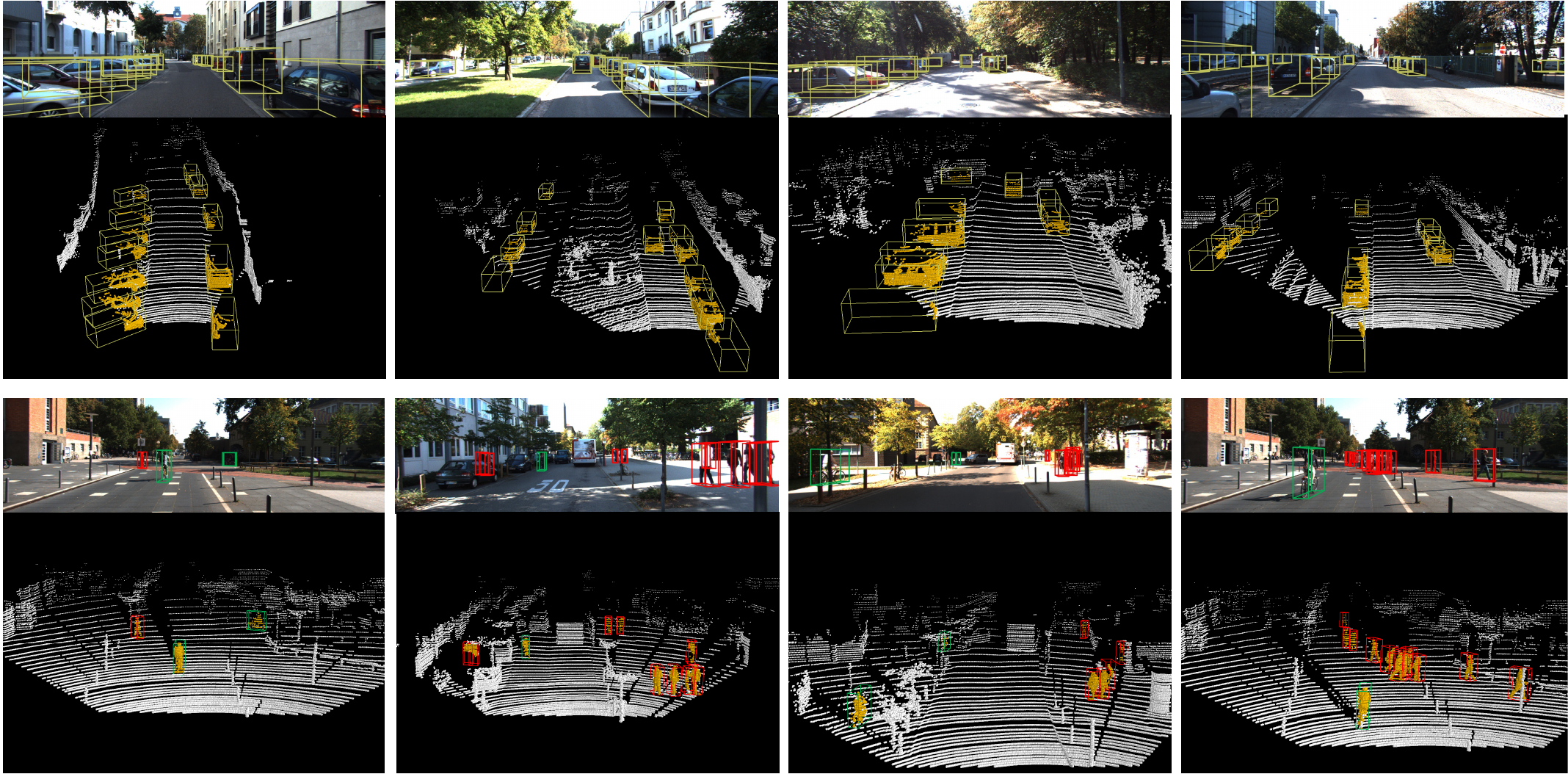}\\
	\end{tabular}
	\caption{Visualization of our results on KITTI test set. Cars, pedestrians and cyclists are highlighted in yellow, red and green respectively. The upper row in each image is the 3D object detection result projected onto the RGB image. The other is the result in the LiDAR phase.}
	\label{fig:results}
\end{figure*}

Compared to LiDAR-only detectors, and voxel or point methods, our method works best on all three classes. Specifically, on Car detection, STD achieves a better AP by $1.87 \%$, $2.64 \%$ and $3.97 \%$ compared to PointRCNN \cite{shi2018pointrcnn}, PointPillars \cite{lang2018pointpillars} and SECOND \cite{yan2018second} respectively on the moderate set. The improvement on the hard set is more significant -- $7.74 \%$, $7.76 \%$ and $9.86 \%$ increase respectively. We present several qualitative results in Figure \ref{fig:results}.

\subsection{Ablation studies}
For ablation studies, we follow VoxelNet \cite{VOXELNET} to split the official training set into a {\it train} set of 3,717 images/scenes and a {\it val} set of 3,769 images/scenes. Images in train/val set belong to different video clips. Following \cite{VOXELNET}, all ablation studies are conducted on the {\tt car} class due to the relatively large amount of data to make system run stably.

\begin{table}[t]
   \centering \addtolength{\tabcolsep}{-1pt}
   \footnotesize
   \begin{tabular}{|c|ccc|}
      \hline
      Class & Easy & Moderate & Hard \\
      \hline
      Car (BEV) & 90.5 & 88.5 & 88.1 \\
      Car (3D) & 89.7 & 79.8 & 79.3 \\
      \hline
      Pedestrian (BEV) & 75.9 & 69.9 & 66.0 \\
      Pedestrian (3D) & 73.9 & 66.6 & 62.9 \\
      \hline
      Cyclist (BEV) & 89.6 & 76.0 & 72.7 \\
      Cyclist (3D) & 88.5 & 72.8 & 67.9 \\
      \hline
   \end{tabular}\vspace{0.1cm}
   \caption{3D and BEV detection AP on KITTI val set.}
   \label{tab:kittival}
\end{table}

\begin{table}[t]
   \centering \addtolength{\tabcolsep}{-1pt}
   \footnotesize
   \begin{tabular}{|c|ccc|}
      \hline
      Method & Easy & Moderate & Hard \\
      \hline
      MV3D \cite{MV3D} & 71.29 & 62.68 & 56.56 \\
      AVOD \cite{AVOD} & 84.41 & 74.44 & 68.65 \\
      VoxelNet \cite{VOXELNET} & 81.97 & 65.46 & 62.85 \\ 
      SECOND \cite{yan2018second} & 87.43 & 76.48 & 69.10 \\
      F-PointNet \cite{FPOINTNET} & 83.76 & 70.92 & 63.65 \\
      PointRCNN \cite{shi2018pointrcnn} & 88.88 & 78.63 & 77.38 \\
      \hline
      Ours (without IoU branch) & 88.8 & 78.7 & 78.2 \\
      Ours (IoU branch) & \bf 89.7 & \bf 79.8 & \bf 79.3 \\
      \hline
   \end{tabular}\vspace{0.1cm}
   \caption{3D detection AP on KITTI val set of our model for ``Car" compared to other state-of-the-art methods.}\vspace{-0.05in}
   \label{tab:kittival_compare}
\end{table}

\vspace{-0.1in}
\paragraph{Results On Validation Set}
We first report the performance on KITTI val set in Table \ref{tab:kittival}. Our results on validation set compared to other methods are listed in Table \ref{tab:kittival_compare}. Unlike voxel-based methods of \cite{VOXELNET} and \cite{yan2018second}, our model preserves more structure and appearance details, leading to better performance. Compared to point-based methods, the proposal generation module and IoU branch keep more accurate proposals and high-quality predictions, which brings higher AP especially on the hard set. 
We compare recall among different 2-stage object detectors in Table \ref{tab:whether_proposal}, demonstrating the powerful of our proposal generation module.

\vspace{-0.1in}
\paragraph{Effect of Anchors Receptive Field}
Given that anchors play an important role, it is important to make anchors cover as much as possible ground-truth area while not consuming much computation. We use spherical receptive field that has only one radius for each detection model. In order to justify the effectiveness of this design, we conduct  experiments varying the shape and size of receptive fields. Average Recall (AR) with IoU threshold 0.7 is the metric. The result is shown in Table \ref{tab:whether_sphere}. First, ``cuboid" shape of receptive field needs more than one angles, \ie $(0, \pi / 2)$, because of the disproportion between length and width, leading to $2 \times$ more data and accordingly more computation. ``Cuboid'' with only one orientation causes $1.5 \%$ decrease in terms of AR. Moreover, spherical receptive field brings additional context information, which benefits anchor classification and regression.

\begin{table}[t]
   \centering \addtolength{\tabcolsep}{-1pt}
   \footnotesize
   \begin{tabular}{|c|c|c|c|}
       \hline
       Methods & Proposals num & IoU threshold & Recall\\
       \hline
       AVOD \cite{AVOD} & 50 & 0.5 & 91.0 \\
       Ours & 50 & 0.5 & \bf 96.3 \\
       \hline
       PointRCNN \cite{shi2018pointrcnn} & 100 & 0.7 & 74.8 \\
       Ours & 100 & 0.7 & \bf 76.8 \\
      \hline
   \end{tabular}\vspace{0.1cm}
   \caption{Recall of proposals on KITTI val set compared to other methods with the same proposal number and IoU threshold.}
   \label{tab:whether_proposal}
\end{table}

\begin{table}[t]
   \centering \addtolength{\tabcolsep}{-1pt}
   \footnotesize
   \begin{tabular}{|c|c|c|c|}
       \hline
       shape & anchors amount & proposals num & Recall(IoU=0.7) \\
       \hline
       cuboid & $1 \times$ & 100 & 74.2 \\
       cuboid & $2 \times$ & 100 & 75.7 \\
       sphere & $1 \times$ & 100 & \bf 76.8 \\
      \hline
   \end{tabular}\vspace{0.1cm}
   \caption{Recall of generated proposals on KITTI val set.}
   \label{tab:whether_sphere}
\end{table}

\begin{table}[t]
   \centering \addtolength{\tabcolsep}{-1pt}
   \footnotesize
   \begin{tabular}{|c|c|c|c|c|c|}
       \hline
       semantic & canonized  & Easy & Moderate & Hard \\
       \hline
       - & - & 38.7 & 31.1 & 26.0 \\
       $\surd$ & - & 82.5 & 67.6 & 67.2 \\
       $\surd$ & $\surd$ & \bf 88.8 & \bf 78.7 & \bf 78.2 \\
      \hline
   \end{tabular}\vspace{0.1cm}
   \caption{3D object detection AP on KITTI val set. A tick in “canonized” item means using canonical coordinates rather than original coordinates as part of feature. A tick in ``semantic" means using points feature from 3D semantic segmentation backbone in proposal feature.}\vspace{-0.05in}
   \label{tab:whether_coord}
\end{table}

\vspace{-0.1in}
\paragraph{Effect of Proposal Feature}
Our proposal features are with canonical coordinates and 3D semantic features. We quantify their benefits using original points coordinates as our baseline. As shown in Table \ref{tab:whether_coord}, using 3D segmentation features results in around $36.5 \%$ performance boost on moderate set in terms of AP. It means global context information enhances model capability greatly. With canonical transformation, AP increases by $11.1 \%$ on moderate set. 

\begin{table}[t]
	\centering \addtolength{\tabcolsep}{-1pt}
	\footnotesize
	\begin{tabular}{|c|c|c|c|c|c|}
		\hline
		NMS & Soft-NMS & 3D & Easy & Moderate & Hard \\
		\hline
		$\surd$ & - & - & 88.8 & 78.7 & 78.2 \\
		- & $\surd$ & - & 88.9 & 79.0 & 78.4 \\
		- & - & $\surd$ & \bf 89.7 & \bf 79.8 & \bf 79.3 \\
		\hline
	\end{tabular}\vspace{0.1cm}
	\caption{3D object detection AP on KITTI val moderate set. Our experiments analyze influence of our 3D IoU branch. ``3D" means using 3D IoU branch for post-processing.}
	\label{tab:whether_iou_branch}
\end{table}

\begin{table}[t]
	\centering \addtolength{\tabcolsep}{-1pt}
	\footnotesize
	\begin{tabular}{|c|c|c|c|}
		\hline
		NMS sorting score & Easy & Moderate & Hard \\
		\hline
		3D-IoU & 89.0 & 79.1 & 78.7 \\
		cls-score $\times$ 3D-IoU & \bf 89.7 & \bf 79.8 & \bf 79.3 \\
		\hline
	\end{tabular}\vspace{0.1cm}
	\caption{3D object detection AP on KITTI val moderate set. Our experiments analyze influence of different ways to use 3D IoU branch. ``3D-IoU'' means only using 3D IoU as NMS sorting score. ``cls-score $\times$ 3D-IoU'' indicates the way we describe in Section \ref{IoU_estimation_framework}.\vspace{-0.1in}}
	\label{tab:way_of_iou_branch}
\end{table}

\vspace{-0.1in}
\paragraph{Effect of IoU Branch}
Our 3D IoU prediction branch estimates the localization quality to finally increase performance. As illustrated in Table \ref{tab:whether_iou_branch}, our 3D IoU guided NMS outperforms traditional methods of NMS and soft-NMS by $1.1 \%$ and $0.8 \%$ on moderate set respectively, manifesting the usefulness of this branch. We note directly taking predicted 3D IoU as the NMS sorting criterion, as shown in Table \ref{tab:way_of_iou_branch}, performs not well. The reason is that only positive proposals are considered in IoU branch, while classification score can tell positive predictions from negative ones. Accordingly, combination of classification score and predicted IoU is very effective.

\vspace{-0.1in}
\paragraph{Inference Time}
The total inference time of STD is 80ms on a TitanV GPU where the PointNet++ backbone takes 54ms, the proposal generation module including PointNet and NMS takes 10ms, PointsPool layer takes about 6ms, and the second stage with two branches takes 10ms. STD is the fastest model among all point-based methods and multi-view methods, manifesting the reasonable design of STD. Note that, we merge batch normalization into convolution layers, and split the input point cloud of first SA level $(16K)$ in PointNet++ to $(32 \times 512)$ for parallel computation so as to shorten inference time, resulting in 25ms and 50ms speedup respectively with performance unchanged.

\section{Conclusion}
We have proposed a new two-stage 3D object detection framework that takes advantages of both voxel- and point-based methods. We introduce spherical anchors based on points and refine them for accurate proposal generation without loss of localization information in the first stage. Then a PointsPool layer is applied for generating compact representations for proposals which is beneficial to reduce inference time. The second stage reduces incorrect removal in post-process which further improves performance. Our model works decently on 3D detection, especially on the hard set. 

{\small
\bibliographystyle{ieee}
\bibliography{egbib}
}

\end{document}